\documentclass[letterpaper, 10 pt, conference]{ieeeconf}  

\IEEEoverridecommandlockouts                              

\usepackage{mathptmx} 
\usepackage{times} 
\usepackage{amsmath} 
\usepackage{amssymb}  

\usepackage{cite}
\usepackage{graphicx} 
\usepackage{float}
\usepackage{bm}
\usepackage{color}
\usepackage{amsmath}
\usepackage{algorithm}
\usepackage[noend]{algpseudocode}

\title{\LARGE \bf
	Trajectory-Optimized Sensing for Active Search of Tissue Abnormalities in Robotic Surgery
}

\author{Hadi Salman$^{1}$, Elif Ayvali$^{1}$, Rangaprasad Arun Srivatsan$^{1}$, Yifei Ma$^{2}$, Nicolas Zevallos$^{1}$, Rashid Yasin$^{3}$, \\Long Wang$^{3}$, Nabil Simaan$^{3}$ and Howie Choset$^{1}$
\thanks{This work was was supported by NRI Large grants IIS-1426655 and IIS-1327566.}
\thanks{$^{1}$H.Salman, E. Ayvali, R. A. Srivatsan, N. Zevallos, and H. Choset are with the Robotics Institute at Carnegie Mellon University, Pittsburgh,PA 15213, USA
	{\tt\small (hadis@andrew., eayvali@, rarunsrivatsan@,  nzevallo@, choset@) cmu.edu}}
\thanks{$^{2}$Y. Ma is with the Machine Learning Department at Carnegie Mellon University, Pittsburgh,PA 15213, USA
	{\tt\small (yifeim@cs.cmu.edu)}}
\thanks{$^{3}$R. Yasin, L. Wang, and N. Simaan are with the Mechanical Engineering Department at Vanderbilt University, Nashville, TN 37235, USA
	{\tt\small ((rashid.m.yasin@,  long.wang@, nabil.simaan@)vanderbilt.edu)}}}

\begin{document}

\maketitle
\thispagestyle{empty}
\pagestyle{empty}

\begin{abstract}
In this work, we develop an approach for guiding robots to automatically localize and find the shapes of tumors and other stiff inclusions present in the anatomy. Our approach uses Gaussian processes to model the stiffness distribution and active learning to direct the palpation path of the robot. The palpation paths are chosen such that they maximize an acquisition  function provided by an active learning algorithm. Our approach provides the flexibility to avoid obstacles in the robot's path, incorporate uncertainties in robot position and sensor measurements, include prior information about location of stiff inclusions while respecting the robot-kinematics. To the best of our knowledge this is the first work in literature that considers all the above conditions while localizing tumors. The proposed framework is evaluated via simulation and experimentation on three different robot platforms: 6-DoF industrial arm, da Vinci Research Kit (dVRK), and the Insertable Robotic Effector Platform (IREP). Results show that our approach can accurately estimate the locations and boundaries of the stiff inclusions while reducing exploration time. 
\end{abstract}

\section{Introduction}
\begin{figure}
	\centering
	\includegraphics[width=.8\linewidth]{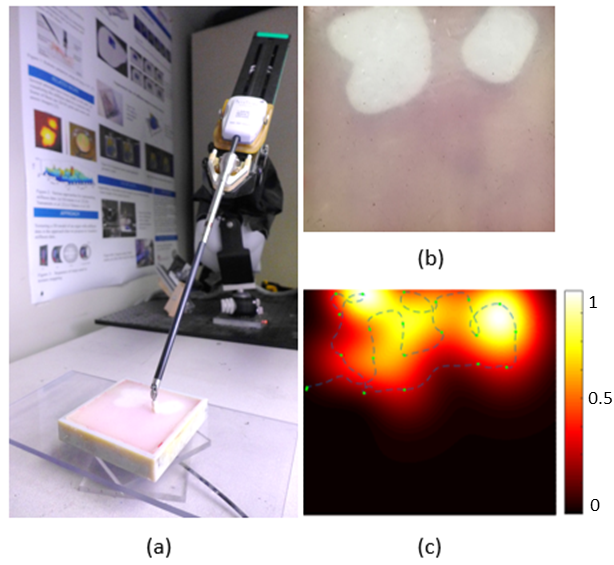}
	\caption{(a) Experimental setup showing da Vinci Research Kit (dVRK), equipped with a spherical tool tip.  (b) Silicone phantom organ with embedded stiff inclusions. (c) Stiffness map as estimated by our approach using active area search and continuous palpation. The estimated stiffness map accurately reveals the location and shape of the two embedded stiff inclusions.}
	\label{fig:davincip}
\end{figure}
Surgeons typically rely on palpation to develop a haptic understanding of the anatomy. They analyze the force and deflection feedback from palpation to localize tumors and sensitive anatomy such as nerve bundles, tendons and arteries. Information from palpation can help surgeons in forming a better understanding of the surgical scene and in achieving a correlation between pre-operative imaging information and the surgical scene. When performing minimally invasive surgery (MIS), often there is a loss of haptic understanding of the anatomy. In order to restore the lost information, several works in literature have focused on developing miniature tactile and force sensors~\cite{puangmali2008state, xu2008investigation, trejos2009robot, gafford2014monolithic, mckinley2015single,li2017development}.

Many groups have looked into using surgical robots for autonomously exploring an organ with discrete probing  motion~\cite{yamamoto2009tissue,beccani2014wireless}, rolling motion~\cite{liu2010rolling} and cycloidal motion~\cite{goldman2013algorithms} to obtain a stiffness map. These works commonly direct the robot along a predefined path that scans the entire organ or region of interest~\cite{howe1995remote,yamamoto2009tissue, beccani2014wireless, srivatsan2016complementary}. Some of these works~\cite{goldman2013algorithms,nichols2015methods} use adaptive grid resolution to increase palpation resolution around boundaries of regions of interest marked by high stiffness gradients~\cite{goldman2013algorithms,nichols2015methods}.
To detect the boundaries of stiff inclusions Nichols~\emph{et al.}\cite{nichols2015methods} used a support vector machine classifier to localize the boundaries of the tumors. Their method, however, requires training the classifier using elastograpy images. 

In order to reduce the exploration time, Bayesian optimization-based approaches have been developed for tumor localization by directing the exploration to stiff regions~\cite{ayvali2016using,chalasani2016concurrent, garg2016tumor,AyvaliRAL2017, preetham2017online}. 
These approaches model tissue stiffness as a distribution defined on the surface of the organ where each point on the surface is associated with a random variable. Bayesian optimization is then used to find the global maxima of the stiffness distribution. The assumption is that finding the global maxima of the stiffness distribution correspond to locating the stiff inclusions. Ayvali~\emph{et al.}~\cite{ayvali2016using} sequentially select the next location to probe the organ, and predict the stiffness distribution and the location of the global maximum after every measurement, while Chalasani~\emph{et al.}~\cite{chalasani2016concurrent} update after collecting several samples over finite time along a trajectory that directs the robot to the high stiffness regions. In a more recent work, Chalasani~\emph{et al.}~\cite{preetham2017online} incrementally estimate local stiffness and geometry while the organ is palpated along predefined trajectories or under telemanipulation. Garg~\emph{et al.}~\cite{garg2016tumor} direct the exploration to areas where the predicted stiffness values are within a percentage of the current estimated maximum to favor locations around the maximum and not just at the expected maximum.

However, none of these approaches explicitly encode the goal of extracting the shape of the stiff inclusion. The only goal that is encoded through a Bayesian optimization framework is to find the global maximum. As a consequence, the robot ends up mainly exploring around high stiffness regions before expanding to the boundary of the inclusion and other regions. Prior works  commonly demonstrate results using a single stiff inclusion (single maximum)~\cite{chalasani2016concurrent,garg2016tumor}. When multiple inclusions are present (multiple global and local maxima) the algorithm is initialized with a coarse grid to ensure exploration of all regions~\cite{ayvali2016using}. 
 
In this work, we present a formulation that leverages state-of-the-art active learning methods as the objective to optimize robot's trajectories and explicitly encodes  search of stiff regions and their boundaries. Compared to the existing works on active learning~\cite{ma2014active,gotovos2013active}, our formulation incorporates constraints due to the robot's motion model, restricts areas in the search domain, and captures uncertainty in the measurements. We show experimental results with a variety of robotic platforms both using discrete probing and along a continuous path that is optimized using stochastic trajectory optimization\footnote{The code base is publicly available at https://github.com/biorobotics/trajectory-optimized-active-search}.


\section{Background}
\label{sec:Background}
\subsection{Gaussian Process Regression}\label{sec:GP}
In our work, we utilize Gaussian processes (GPs) to model the distribution of stiffness on the organ. GP is a popular tool used to perform nonparameteric regression. Intuitively a GP can be viewed as a distribution over functions. By using GP, we assume a smooth change in the stiffness distribution across the organ. Since every point on the organs's surface can be mapped in a 2D grid, the domain of search used is $X\subset {\rm I\!R^2}$. The measured force and position after probing the organ by the robot at $\bm{x}$ provides the stiffness estimation represented by $y$.

A GP is defined by its mean and covariance functions $f_{GP}$ and $k$ respectively. Given a $d$-dimensional search domain $X\subset {\rm I\!R^d}$, the distribution of function values at a point $\bm{x} \in X$ is represented by a random variable, $y$, and has a Gaussian distribution, ${N}(f_{GP}(\bm{x}),\sigma^2(\bm{x}))$ where we abbreviate \mbox{$\sigma^2(\bm{x}) = k(\bm{x},\bm{x})$}.  
Given a set of $n$  observations \mbox{$\bm{\bar{y}}=[y_1, y_2, \dots, y_n]^T$} at \mbox{$\bar{ X}=[\bm{x}_1, \bm{x}_2, \dots, \bm{x}_n]^T$},  GP regression can be used to make predictions on the distribution of  function values at a new point $\bm{x}_* \in X$
\begin{equation}\nonumber
p(y_*|\bm{\bar{y}}) \sim {N}(\bm{K_}*\bm{K}^{-1}\bm{\bar{y}},k_{**}-\bm{K_}*\bm{K}^{-1}\bm{K^T_}*),
\label{eq:GP_eqn}
\end{equation}
where $\bm{K}$ is the $n \times n$ covariance matrix whose elements \mbox{$\bm{K}_{ij}$  $(i, j \in [1,\dots,n])$} are calculated using any positive definite covariance function $k(\bm{x}_i,\bm{x}_j)$ (in this paper we use the squared exponential covariance function). Similarly, $\bm{K}_*$ is a $1 \times n$ vector defined as \mbox{ $\bm{K}_*=[k(\bm{x}_*,\bm{x}_1),\dots,k(\bm{x}_*,\bm{x}_n)]$}, and finally \mbox{$k_{**}=k(\bm{x}_*,\bm{x}_*)$}.

In order to incorporate input uncertainty into the GP, we adopt the formulation of Girard \emph{et al.}~\cite{girard2003learning} that ``corrects'' the covariance function $k(\bm{x},\bm{x}) $ of the GP. This is useful when the uncertainty in the robots position is significant.

\subsection{Active Learning and Bayesian Optimization}\label{sec:background:active_learning}
In many learning scenarios, unlabeled data are plentiful and manually labeling them is expensive. The role of active learning algorithms is to efficiently find which data to label. In this work, the search space is the surface of the organ and labeling data corresponds to assigning a binary value to every point on the organ's surface: normal tissue vs. tissue abnormality. We consider in this work various active learning algorithms: Active area search (AAS), active level sets estimation (LSE), and uncertainty sampling (UNC), and compare them with Bayesian optimization algorithm, which gained interest in recent works.
\subsubsection{Active Area Search}
This algorithm discretizes the search domain into a set of regions G = $\{g_1,g_2,\dots,g_N\}~\subset~X$ and classifies each as region-of-interest (tissue abnormalities corresponding to regions that have high stiffness) if the average estimated latent function (stiffness function in our case) is above some threshold $\tau$ with high probability $\theta$. AAS sequentially queries at a point $\bm{x}_*$ that maximizes the expected sum of binary rewards $r_g$ defined over each region $g\in G$ as,
\begin{equation}\nonumber
r_g	=
\begin{cases}
1, & \text{if}\ p\left(f_g>\tau |\bar X,\bm{\bar{y}},(\bm{x}_*,y_*)\right)>\theta \\
0, & \text{otherwise}
\end{cases}
\end{equation}
where $y_*$ is the observation at $\bm x_*$ and $f_g$ is the average area integral of $f_{GP}$ over the region $g$ and is defined as,
\begin{equation}\nonumber
	f_g(\bm{x}) = \frac{1}{A_g}\int_g f_{GP}(\bm{x})d\bm{x}
\end{equation}
where $A_g$ is the area of $g$. Thus, AAS sequentially samples the point $\bm x_*$ that maximizes the expected total reward, i.e.,
\begin{align}\label{eq:aasaqfxn}
\bm x_* =&\operatorname*{arg\,max}_{\bm x\in X}\sum_g \mathbb{E}[r_g|\bar X,\bm{\bar{y}},(\bm{x},y)].
\end{align}
For more details, we refer the reader to the work of \mbox{Ma~\emph{et al.}~\cite{ma2014active}}. 

\subsubsection{Active Level Set Estimation}
This algorithm determines the set of points, for which an unknown function (stiffness map in our case) takes value above or below some given threshold level $h$. LSE guides both sampling and classification based on GP-derived confidence
bounds. The mean and covariance of the GP can be used to define a confidence interval, 
\begin{equation}\nonumber
Q_t(\bm x) = \left[f_{GP_t}(\bm x) \pm \beta^{1/2} \sigma_t(\bm x)\right]
\end{equation} 
for each point $\bm x \in \bar X$, where the subscript $t$ refers to time. Furthermore, a confidence region $C_t$ which results from intersecting successive
confidence intervals can be defined as,
\begin{equation}\nonumber
C_t(\bm x) = \bigcap_{i=1}^t Q_i(\bm x)
\end{equation}
LSE then defines a measure of classification ambiguity $a_t(\bm x)$ defined as,
\begin{equation}
\label{eq:lseaqfxn}
a_t(\bm x) = \min\left\{\max(C_t(\bm x))-h, h - \min(C_t(\bm x)) \right\}
\end{equation}
LSE chooses sequentially queries (probes) at $\bm x_*$ such that,
\begin{equation}\nonumber
\bm x_* = \operatorname*{arg\,max}_{\bm x\in X} {a_t(\bm x)}.
\end{equation}
For details and how to select the parameter $h$, we refer the reader to the work of Gotovos~\emph{et al.}~\cite{gotovos2013active}. 

\subsubsection{Uncertainty Sampling}
The Uncertainty Sampling (UNC) algorithm explores locations that have high marginal variance in the GP posterior distribution \cite{seo2000gaussian}. The samples sequentially picked by UNC are blind to the outcome of the search.

\subsubsection{Bayesian Optimization}
In addition to the above active learning algorithms, we consider Bayesian optimization algorithm (BOA) which has gained recent popularity in the community for haptic exploration. BOA is a sequential sampling strategy for finding the global maxima of black-box functions~\cite{brochu2010tutorial}.  A GP is used as a surrogate for the function to be optimized. BOA uses the posterior  mean, $\mu(\bm{x})$, and variance, $\sigma^2(\bm{x})$, of the GP for all $\bm{x} \in X$, to sequentially select the next best sample as the point that maximizes an objective function such as expected improvement (EI) given by~\cite{brochu2010tutorial}
\begin{align}\nonumber
\bm x_* &= \operatorname*{arg\,max}_{\bm x\in X} {EI(\bm x)}\\
EI(\bm{x})&=
\begin{cases}
(f_{GP}(\bm{x})-y^+) \Phi(z)+\sigma(\bm{x}) \phi (z)     & \text{if }  \sigma(\bm{x})>0    \\
0        & \text{if }  \sigma(\bm{x})=0
\end{cases}
\label{eq:EI}
\end{align}
where \mbox{$z=\left(\frac{f_{GP}(\bm{x})-y^+}{\sigma(\bm{x})}\right)$}, $y^+$ is the current maximum. $\phi(\cdot)$ and $\Phi(\cdot)$ are the probability density function and cumulative distribution function of the standard normal distribution, respectively.

\subsection{Cross-Entropy Method}
\label{CE-method}



The coss-entropy (CE) method is a general optimization framework that was used in \cite{kobilarov2012cross} for trajectory optimization of nonlinear dynamic systems. The CE method treats an optimization problem as an estimation problem of rare-event probabilities. The rare event of interest in a trajectory optimization framework is to find a parameter $\bm z$ (corresponding to a parametrization of the trajectory) whose cost $J(\bm z)$ is very close to the cost of an optimal parameter $\bm z^*$. It is assumed that the parameter $\bm z \in Z$ is sampled from a Gaussian mixture model defined as
\begin{equation}\label{eq:GMM}
p(\bm z;v)=\sum_{k=1}^{K}\frac{w_k}{\sqrt{(2\pi)^{n_{z}} |\Sigma_k|}} e^{-\frac{1}{2}(\bm z-\mu_k)^T \Sigma_k^{-1}(\bm z-\mu_k)}
\end{equation}
where $v=(\mu_1,\Sigma_1,...,\mu_K,\Sigma_K,w_1,...,w_K)$ corresponds to $K$ mixture components with means $\mu_k$, covariance matrices $\Sigma_k$, and weights $w_k$, where $\sum_{k=1}^{K}w_k=1$. 

The CE method involves an iterative procedure where each iteration has two steps: \textit{(i)} select a set of parameterized trajectories from $p(\bm z;v)$ using importance sampling~\cite{rubinstein2013cross} and evaluate the cost function $J(\bm z)$, (ii) use a subset of \textit{elite} trajectories\footnote{A fraction of the sampled trajectories with the best costs form an elite set. See~\cite{rubinstein2013cross} for details.} and update $v$ using expectation maximization~\cite{mclachlan2004finite}. After a finite number of iterations $p(\bm z;v)$ approaches to a delta distribution, thus the sampled trajectories remains unchanged. For implementation details, the reader is referred to~\cite{de2005tutorial}.

\section{The Active Search Framework}
\label{sec:Stochastic_Active_Learning}
We pose the active search problem as a constrained optimization problem subject to constraints associated with the motion model of the robot. We then demonstrate that obstacle avoidance can be easily incorporated into this framework by penalizing the sampled trajectories that collide with arbitrarily-shaped obstacles or that pass through restricted regions.

\subsection{Stochastic Trajectory Optimization}
\label{trajopt} 
We use a trajectory optimization framework that allows the search of tumors to be done by any robot with a defined set of motion primitives. 
Consider a robot whose motion model is described by the function  $g\colon Q \times U \to TQ $, such that
\begin{equation}
\label{eq:dynamics}
\dot{\bm{q}}(t)=g(\bm{q}(t),\bm{u}(t)) 
\end{equation}
where $\bm q \in Q$ is the state of the robot and $\bm u \in U$ denotes the set of controls to the robot.


The goal of a trajectory optimizer is to compute the optimal controls $\bm{u}^*$ over a time horizon $t \in (0,t_f]$ that  minimize a cost function such that,
\begin{equation}
\label{eq:CEM}
\begin{gathered}
\bm{u}^*(t)=\operatorname*{arg\,min}_{\bm{u}}\int_0^{t_f}  C(\bm{u}(t),\bm{q}(t))dt,   \\  
\mbox{subject to } \dot{\bm{q}}(t)=g(\bm{q}(t),\bm{u}(t)),\\
 F(\bm{q}(t))\geq 0 , \\
\bm{q}(0)=\bm{q}_0 ,
\end{gathered}
\end{equation}
where $\bm{q_0}$ is the initial state af the robot and $C\colon U \times Q \to \mathbb{R} $ is a given cost function, and $F$ describes the constraints such as joint limits and obstacles in the environment  

Following the notation in~\cite{kobilarov2012cross},
a trajectory defined by the controls and states over the time
interval $[0,T]$ is denoted by the function \mbox {$\pi \colon [0,T] \to U \times Q$}, i.e. $\pi(t)=( \bm{u}(t) , \bm{q}(t))$ for all $t \in [0, T].$ The space of all trajectories originating at $\bm{q}_0$ and satisfying Eq.~(\ref{eq:dynamics}) is given by
\begin{equation}\nonumber
\begin{aligned}
P=\{ \pi \colon t \in [0, T] \to \left( \bm{u}(t),\bm{q}(t)\right)| 
\dot{\bm{q}}(t)=g(\bm{q}(t),\bm{u}(t)), \\
 \bm{q}(0)=\bm{q}_0, T>0 .\} 
 \end{aligned}
\end{equation}
Let us consider a finite-dimensional parameterization of trajectories in terms of vectors $\bm{z} \in Z$ where $Z\subset \mathbb{R}^{n_z}$ is the parameter space. Let us assume that the parameterization is given by a function $\varphi:Z \to P$ according to $\pi=\varphi(z)$. The $(\bm{u},\bm{q})$ tuples along a trajectory parameterized by $\bm{z}$ are written as $\pi(t)=\varphi(\bm{z},t)$. One choice of parameterization is to use motion primitives defined as \mbox{$\bm{z} = (\bm{u}_1, \tau_1,...,\bm{u}_j, \tau_j)$} where each $\bm u_i$, for $1 \leq i \leq j$, is a constant control input applied for duration $\tau_i$. 


In this work, we use a Dubins car model for modeling the motion of the robot. This model generates intuitive paths composed of straight line segments and circles similar to the palpation motion physicians use. For a Dubins car model whose motion is restricted to a plane we can represent its trajectories as a set of connected motion primitives consisting of either straight lines with constant velocity $v$ or arcs of radius $v/w$ where $w$ is the turning rate. A primitive can be defined by a constant controls ($v, w$). The duration of each primitive is constant and $\tau > 0$. The trajectory of the robot can be parameterized using $m$ primitives, and this finite dimensional parameterization is represented by  a vector $\bm{z} \in \mathbb{R}^{2m}$ such that,
\begin{equation}\nonumber
\bm{z} = (v_1,w_1, \dots, v_m, w_m) 
\end{equation}

Now, we can define a cost function, $J \colon Z \to \mathbb{R}$ , in terms of the trajectory parameters as
\begin{equation}
\label{eq:CEM_costfunc}
\begin{gathered}
 J(\bm{z})=\int_0^T C(\varphi(\bm{z},t)) dt
\end{gathered}
\end{equation}
Eq.~(\ref{eq:CEM}) can be restated as finding the optimal $(\bm{u}^*,\bm{q}^*)=\varphi(\bm{z}^*)$ such that
\begin{equation}
\label{eq:CEM_cost}
\bm{z}^*= \operatorname*{arg\,min}_{\bm{z} \in Z_{con}} J(\bm{z}).
\end{equation}
where the constrained parameter space $Z_{con} \subset Z$ is
the set of parameters that satisfy the boundary conditions and
constraints in Eq.~(\ref{eq:CEM}). 

We then employ the cross entropy (CE) method to optimize the parameters of the trajectory as described in Section \ref{CE-method}. There are other sampling-based global optimization methods such as Bayesian optimization~\cite{snoek2012practical}, simulated annealing~\cite{van1987simulated}, and other variants of stochastic optimization~\cite{zhigljavsky2007stochastic} that can also be used to optimize parameterized trajectories. 
We use the CE method because it  utilizes importance sampling to efficiently compute trajectories that have lower costs after few iterations of the algorithm, and it has been shown to perform well for trajectory optimization of nonlinear dynamic systems~\cite{kobilarov2012cross}.

\begin{algorithm}[t]
	\caption{Discrete Palpation}
	\label{pseudo_discrete}
	\begin{algorithmic}[1]
		\State Initialize the GP with zero mean and squared exponential  covariance function
		\State $\bm x^*\gets random $ \Comment random initialization of probed point
		\While{TRUE}
		\State Palpate at $\bm x^*$
		\State Calculate stiffness at probed points
		\State	{Update GP using the stiffness estimate}
		\State Update acquisition function $\xi_{acq}$ using GP
		\State $\bm x^* \gets \arg \max \xi_{acq}$
		\EndWhile
	\end{algorithmic}
\end{algorithm}

\begin{algorithm}[t]
	\caption{Trajectory-Optimized Continuous Palpation}
	\label{pseudo_continuous}
	\begin{algorithmic}[1]
		\State Initialize the GP with zero mean and squared exponential  covariance function
		\State $\bm{z}^*\gets random $ ~~~~\Comment random initial trajectory in the \par ~~~~~~~~~~~~~~~~~~~~~~ space of motion primitives 
		\While{TRUE}
		\State {Execute trajectory $\bm{z}^*$}
		\State Collect stiffness measurements along the trajectory
		\State{Update GP using the stiffness estimate}
		\State Update acquisition function $\xi_{total}$ using GP 
		\Comment (\ref{eq:total_acqusition}) 
		\State $\bm{z}^* \gets \operatorname*{arg\,min}_{\bm{z} \in Z_{con}} J(\bm{z})$
		\Comment (\ref{eq:cost_of_acq})
		\EndWhile
	\end{algorithmic}
\end{algorithm}


\subsection{Objective Function for Active Search}

In this section, we introduce the objective function that we optimize for in the stochastic trajectory optimization framework presented in Section~\ref{trajopt}. 

The problem of finding the location and shape of the stiff inclusions while considering various inherent constraints can be modeled as an optimization problem. However, an exact functional form for such an optimization is not available in reality.Hence, we maintain a probabilistic belief about the stiffness distribution and define a so called ``acquisition function'' to determine where to sample next.

The trajectory optimization problem posed in Eq. (\ref{eq:CEM_cost}) can be solved by defining the cost function $J(\bm{z})$ as,
\begin{equation}\label{eq:cost_of_acq}
J(\bm{z}) = -\int_{\phi(\bm{z})} \xi_{total}(\bm{q})d\bm{q}  
\end{equation}
where $\bm{z}$ is sampled from a Gaussian mixture model defined in Eq. (\ref{eq:GMM}), $\phi(\bm{z})$ is the sampled trajectory that is parameterized by the motion primitive $\bm{z}$,  and $\xi_{total}$ is total acquisition that is to be maximized by each sample trajectory, and is defined by,
\begin{equation}\label{eq:total_acqusition}
\xi_{total}(\bm{q}) = \eta\left(\xi_{acq}(\bm{q}) + \alpha(t) \xi_{prior}(\bm{q}) \right) 
\end{equation}
where \mbox{$\xi_{acq}$} is a normalized acquisition function defined by any one of the active learning algorithms described in Section \ref{sec:background:active_learning}.
\mbox{$\xi_{acq}$} is defined as the expected total reward from Eq.~\ref{eq:aasaqfxn} when using AAS, the ambiguity  $a_t$ from Eq.~\ref{eq:lseaqfxn} in the case of LSE, the uncertainty in the GP posterior distribution when using UNC
\footnote{This uncertainty associated to the estimated stiffness map by the GP and should not to be confused by the uncertainty in the robot's position or force measurement.}
, and the EI as shown in Eq.~\ref{eq:EI} in the case of BOA.

$\xi_{prior}(\bm{q})$ is a normalized distribution capturing the prior on the locations of the tumors, and it decays as search progresses by means of a decay function $\alpha(t)$. Note that the effect of this term has been studied in detail in our previous work~\cite{ayvali2017utility}. In this work we focus on the effect of $\xi_{acq}$. 
\begin{figure}[t]
	\centering
	\includegraphics[width=.9\linewidth]{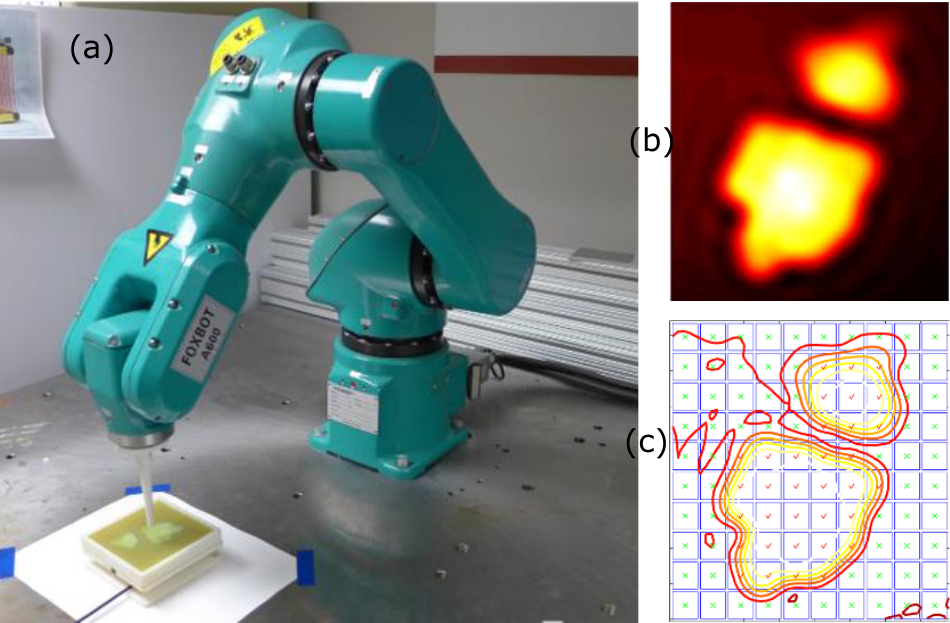}
	\caption{(a) 6 DoF industrial robot arm with a force sensor attached to its end effector. (b) Ground truth stiffness map generated by densely probing a silicone phantom organ. (c) A contour map showing various stiffness levels.}
	\label{fig:croppedfoxbot}
\end{figure}

\subsection{Obstacle Avoidance}
In some surgical scenarios, one may want to avoid palpating certain regions of the organ's surface such as a bony region or regions occupied by other instruments etc. In order to handle such scenarios, our framework can also account for obstacles while searching. Suppose that the search domain $X$ contains $l$ obstacles denoted by $O_1,..., O_l \subset X$. We assume that the robot at state $\bm{q}$ is occupying a region $A(\bm{q})\subset X$. Borrowing the notation in \cite{kobilarov2012cross}, let the function \textbf{prox}($A_1,A_2$) return the closest Euclidean distance between two sets $A_{1,2} \subset X$. This function returns a negative value if the two sets intersect. Therefore, for an agent to avoid the obstacles $O_1,...,O_l$, we impose a constraint of the form shown in Eq.~(\ref{eq:CEM}) expressed as,
\begin{equation}\label{eq:constraint}
F(\bm{q}(t)) = \min_i \mbox{\textbf{prox}}(A(\bm{q}(t)),O_i), ~ \forall t\in[0,\infty).
\end{equation}

\section{Simulation Results}
\label{sec:Sim_Results}
\subsection{Discrete Probing}

\begin{figure}[t]
	\centering
	\includegraphics[width=0.8\linewidth]{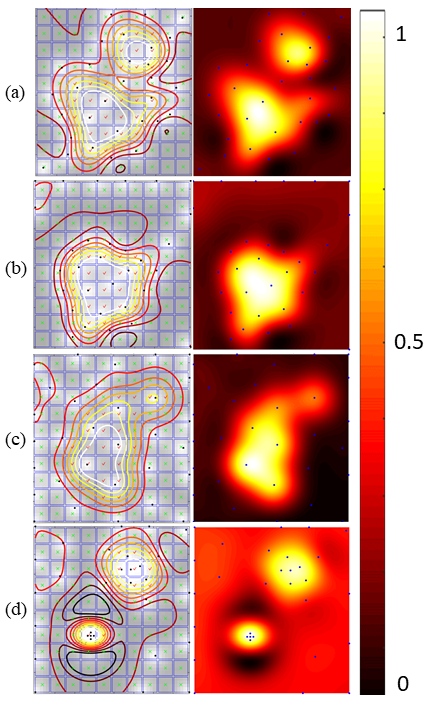}
	\caption{(a) Active area Search (AAS), (b) Active level sets (LSE), (c) Uncertainty sampling (UNC), and (d) Expected improvement (EI) in a tumor search using only 30 probed points. We discretize the search space into regions (squares) as shown in the figures to the left: regions whose average estimated stiffness is above a certain threshold are marked as with tumor and are marked with a red tick. Otherwise, the regions are marked with green cross signifying normal tissue regions.}
	\label{fig:croppedcomparediscreteprobed}
\end{figure}

\begin{figure}[t]
	\centering
	\includegraphics[width=1.0\linewidth]{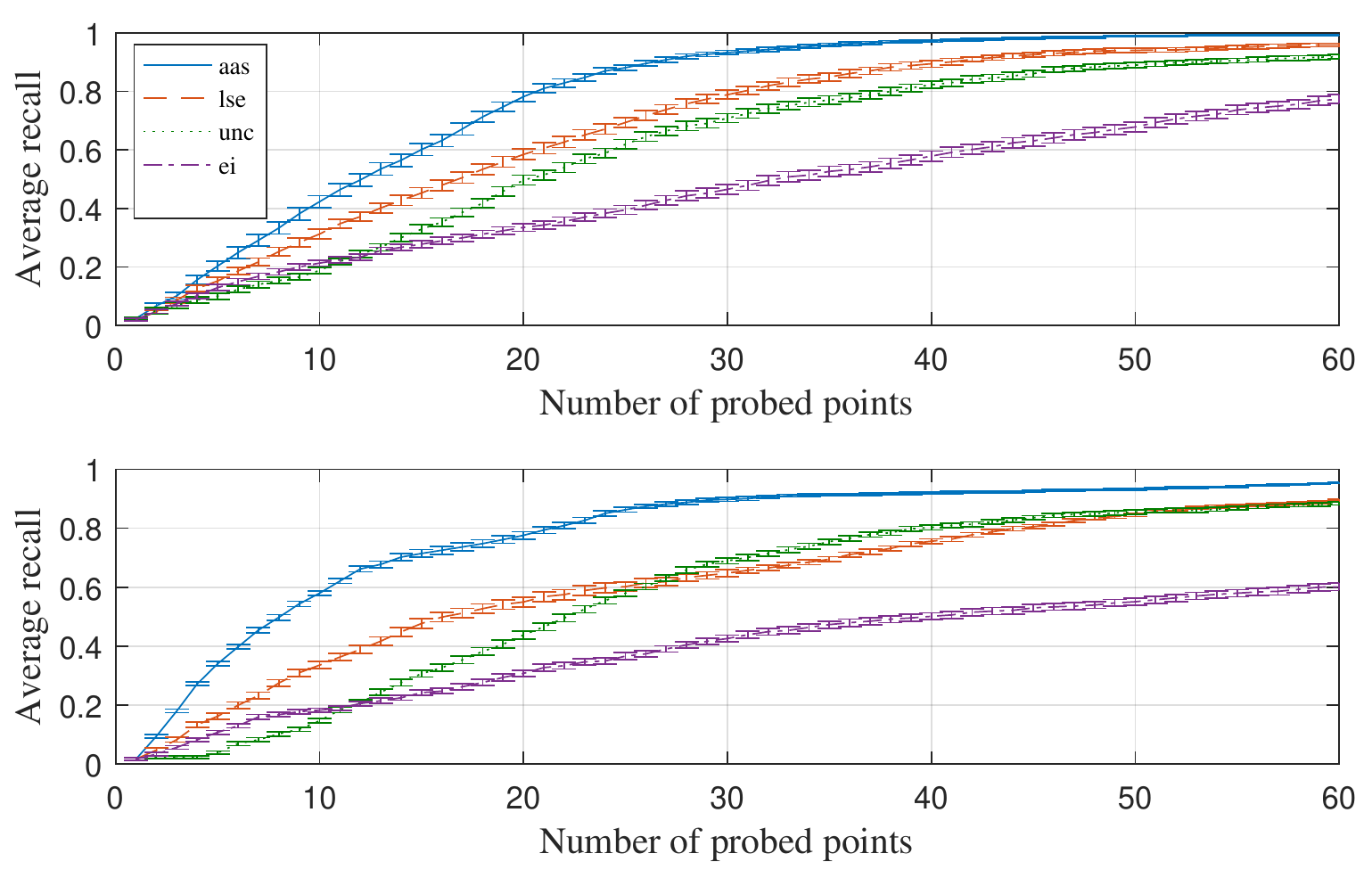}
	\caption{The top plot shows the average recall as a function of the number of the palpated points repeated over 100 simulations of discrete palpation (Algorithm \ref{pseudo_discrete}) with randomly generated ground truths for the stiffness map. The mean and the covariance of each of the four algorithms considered in this paper is reported. The bottom plot shows the average recall for 100 repeated simulations for tumor localization over a fixed ground truth but with random starting probing point for the algorithms.}
	\label{fig:croppedrecall15runs60queries}
\end{figure}

We start by comparing four different sequential probing algorithms which we adopt in this paper as efficient methods to guide our tumor search. This section considers discrete probing scenarios, that are described in Algorithm \ref{pseudo_discrete}, where it is assumed that the robot can reach any point in the search domain.

The robot has no prior knowledge of the locations of the stiff regions. It starts probing at a random location in the 2-D domain. The robot then sequentially decides where to go and probe next such that the acquisition function $\xi_{total}$ associated with each algorithm is maximized. For example, for the active level set estimation (LSE) algorithm, the robot chooses the point with the highest ambiguity in its classification at each step and goes and probes there.
We test the four different algorithms listed in Section \ref{sec:background:active_learning} in a simulated experiment. We use a ground truth of a silicone phantom organ obtained by doing a raster scan using a 6 DoF industrial robot as shown in Fig. \ref{fig:croppedfoxbot}. The results of this simulation are shown in Fig. \ref{fig:croppedcomparediscreteprobed}. We will discuss the results of this experiment in Section \ref{sec:discussion}.

We repeat this experiment 100 times with same parameters of the GP but with randomly generated ground truths. Then we repeat 100 simulations on a fixed ground truth but with random initial probed points. The average recall\footnote{We report the recall since it is a suitable performance measure for regions-of-interest detection problems. The recall is widely used in the Machine Learning community as a performance measure for similar problems.} as a function of the number of probed points for the different algorithms is reported in Fig. \ref{fig:croppedrecall15runs60queries}.
\subsection{Continuous Probing}
Discrete probing does not impose a constraint on the next location to be probed. A robot may not be able to reach the next desired point due to motion constraints. Further, the robot can benefit from collecting information along an optimized path to improve the predictions of tumor location and boundaries.  

\begin{figure}[t]
	\centering
	\includegraphics[width=.8\linewidth]{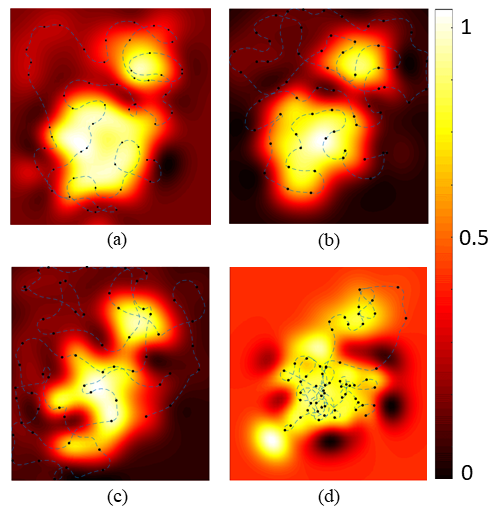}
	\caption{Continuous palpation for tumor localization using (a) AAS, (b) LSE, (c) UNC, and (d) EI acquisition functions in a trajectory optimized search framework represented by Algorithm \ref{pseudo_continuous}.}
	\label{fig:croppedcontinuous-comparison}
\end{figure}

We perform continuous palpation experiments in simulation on the same dataset used in the previous section and shown in Fig. \ref{fig:croppedfoxbot}. The results are shown in Fig. \ref{fig:croppedcontinuous-comparison} and discussed in Section \ref{sec:discussion}. We repeat this experiment 100 times with same parameters of the GP but with randomly generated ground truths. Then we repeat 100 simulations on a fixed ground truth but with random initial starting positions. The average recall as a function of the number of fixed-frequency-sampled measurements along the palpation path for the different algorithms is reported in Fig.~\ref{fig:croppedcontinuousrecallplots}.

 \begin{figure}
 	\centering
 	\includegraphics[width=0.9\linewidth]{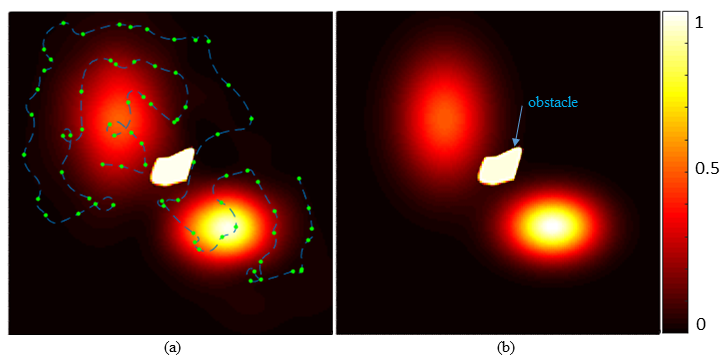}
 	\caption{(a) The trajectory of the robot overlaid on the predicted stiffness function of the search domain. (b) The ground truth of the stiffness function. The white region is an obstacle. The green points along the trajectory are the points which we used to update the GP (probed points). The AAS algorithm is used.}
 	\label{fig:obstaclessnapshot}
 \end{figure}
 
\begin{figure}
	\centering
	\includegraphics[width=1.0\linewidth]{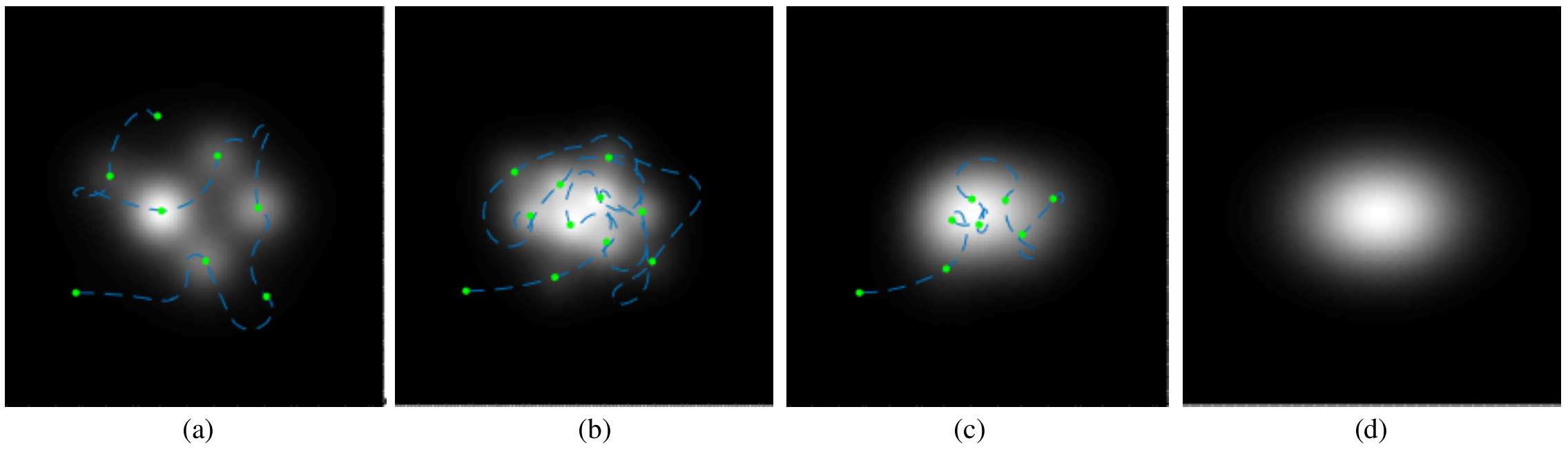}
	\caption{Results showing the estimated stiffness map using Algorithm \ref{pseudo_continuous} with (a) High uncertainty, (b) low uncertainty, and (c) no uncertainty in the robot's position. These uncertainties are propagated through GP according to the formulation in Section \ref{sec:GP}. (d) shows the ground truth of the stiffness map.}
	\label{fig:GP}
\end{figure}

\begin{figure}[h]
	\centering
	\includegraphics[width=1.0\linewidth]{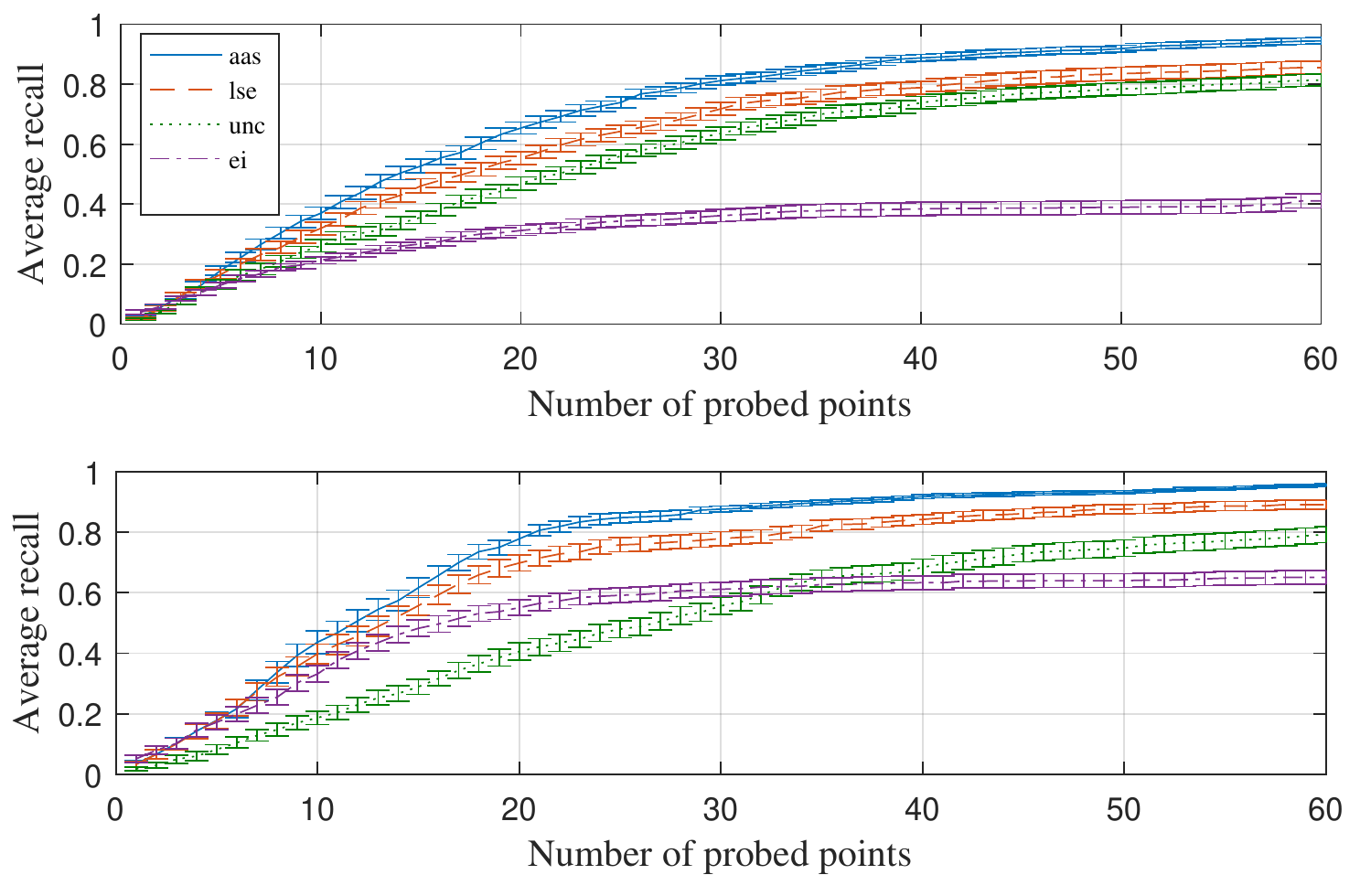}
	\caption{The top plot shows the average recall as a function of the number of the palpated points repeated over 100 simulations of continuous palpation (Algorithm \ref{pseudo_continuous}) with randomly generated ground truths for the stiffness map. The mean and the covariance of each of the four algorithms considered in this paper is reported. The bottom plot shows the average recall for 100 repeated simulations for tumor localization over a fixed ground truth but with random starting probing point for the algorithms.}
	\label{fig:croppedcontinuousrecallplots}
\end{figure}

\subsection{Discussion}\label{sec:discussion}
The BOA is designed to focus on finding the global maxima of a function. Therefore, once a point with high stiffness is detected, the algorithm collects more samples around it instead of moving out from that region and discerning the boundary. This is clearly observed in both Fig~\ref{fig:croppedcomparediscreteprobed}-d as well as Fig.~\ref{fig:croppedcontinuous-comparison}-d. 

LSE approach is designed to improve classification around an implicitly defined level set (defined as a percentage of the maximum estimated stiffness value so far) corresponding to tumor boundaries and as a result can find the boundaries of the tumors fairly well. However,  when each tumor boundary lies on a different level set, the algorithm may spend too much time finding one boundary instead of exploring for other tumors. This is evident from Fig.~\ref{fig:croppedcomparediscreteprobed}(b), where the shape of one tumor is estimated correctly, but in the given number of probings, the second tumor was not detected. 

AAS provides a good balance between finding the boundaries of the tumor and finding the location of multiple tumors as the algorithm searches all areas where the average of the unknown function (stiffness distribution) over the region exceeds the implicit threshold. Both in the case of discrete as well as continuous palpation, the AAS outperforms all the other approaches. The UNC approach has the worst performance since the algorithm is blind to the value of the predicted stiffness distribution.

Fig. 6 shows simulation results for a case where there are restricted regions in the domain that should be avoided. The trajectory planner, using the AAS algorithm, succeeds in avoiding the obstacle while still localizing both tumors.

In scenarios where the there is significant uncertainties in the robot's position, the GP estimate of the stiffness map is affected. This can be taken into account by incorporating this uncertainty in the robot's position as input uncertainty to the GP during training as mentioned in Section \ref{sec:GP}. We simulate in Fig. \ref{fig:GP} three scenarios of tumor search with  different levels of uncertainty in the robot's position. The results shows that as the input uncertainty increases, the estimate of the stiffness map deteriorates.

\section{Experimentation}
\label{sec:experiments}
We validate our results by performing experiments on three different robots (6 DoF industrial robot, dVRK, and IREP) to do autonomous palpation and search for tumors. The experimental studies are all performed on phantom silicone organs with embedded stiff inclusions.  We observe that continuous palpation using AAS produces best results as previously observed in simulation. However, due to space limitations, we only present the results for AAS in all the robot experiments.

\subsection{6-DoF Industrial Arm}
We use a 6-DoF industrial arm as a platform to verify our simulation results and show that our framework runs real-time (See Fig.~\ref{fig:croppedfoxbot}). A commercial force sensor, ATI Nano25 F/T, was attached at the end effector of the robot. As the robot is commanded to move along a trajectory, we continuously collect force measurements from the sensor and position measurements from the kinematics. We employ a linear stiffness model and use the slope of the line that best fits the force-displacement profile similar to~\cite{srivatsan2016complementary} to find a scalar stiffness value at every location on the organ.

\begin{figure}
	\centering
	\includegraphics[width=0.9\linewidth]{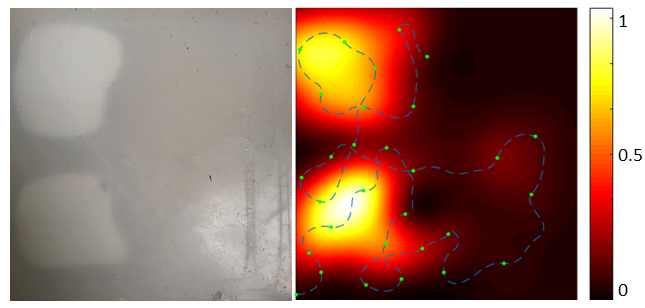}
	\caption{Result of the experiments performed using the 6-DoF Industrial Arm. Left: Top view of the silicon phantom organ showing two stiff inclusions. Right: Stiffness map as estimated by our approach. The palpation trajectory is superimposed on the stiffness map.}
	\label{fig:foxbotsnapshot}
\end{figure}

Fig.~\ref{fig:foxbotsnapshot} shows the stiffness map as estimated by using our framework to palpate the organ along a continuous trajectory. The estimated stiffness map clearly reveals the location and the shape of the stiff inclusions without wasting time exploring the softer regions of the organ. 

\subsection{da Vinci Research Kit}
We use the open source da Vinci Research Kit (dVRK)~\cite{kazanzides2014open} for evaluating our approach on silicone tissue samples. The dVRK serves as a realistic surgical platform for evaluating the efficacy of tumor search algorithms. In order to perform palpation, we attach a custom 3D printed spherical-head tip to the 8mm needle driver tool of the robot. The silicone tissue sample with embedded stiff inclusions (see Fig.~\ref{fig:davincip}) is placed on top of an ATI Nano25 F/T sensor. Fig.~\ref{fig:davincip} shows the stiffness map as estimated by our approach as well as the superimposed palpation trajectory. The stiffness map accurately reveals the stiff inclusions without wasting time exploring the softer regions in the bottom half of the tissue sample.

\subsection{Insertable Robotic Effector Platform (IREP)}
\begin{figure}[t]
	\centering
	\includegraphics[width=.7\linewidth]{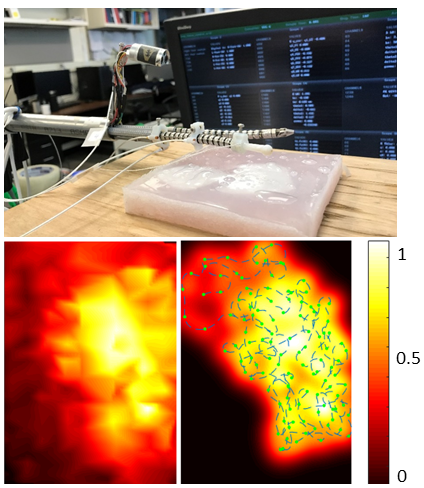}
	\caption{(a) Experimental setup consisting of an Insertable Robotic Effector Platform (IREP), probing a silicone phantom organ which is placed on top of a force sensor. (b) Ground truth stiffness map generated by densely probing the organ. (c) Stiffness map as estimated by continuous palpation using active areas search (AAS). The estimated stiffness map confirms well with the ground truth.Insertable Robotic Effector Platform (IREP).}
	\label{fig:irep}
\end{figure}

The IREP is a two-segment, four-backbone continuum robot actuated with push-pull nitinol wires designed for single port access surgery~\cite{bajo2012integration}. The IREP has an architecture which is very different from conventional rigid link robots and hence provides a challenging platform to demonstrate our approach. The experimental set up is similar to the one used with dVRK and is shown in Fig.~\ref{fig:irep}. While this type of robot architecture is compatible with intrinsic force sensing as in~\cite{xu2008investigation}, the integration of trajectory optimization with intrinsic force sensing on the IREP and the accompanying challenges of uncertainty estimation of pose and force are part of ongoing research. 

Fig.~\ref{fig:irep}(b) shows the ground truth stiffness map as generated by densely probing the organ surface using the IREP over a grid of 330 points. Fig.~\ref{fig:irep}(c) shows the stiffness map as estimated by our approach as well as the superimposed palpation trajectory. In this experiment, we do not perform continuous palpation with the robot, but instead use the data obtained by densely probing with IREP to simulate continuous palpation. The estimated stiffness map qualitatively confirms with the ground truth stiffness map.

\section{Discussions and Future Work}
This work introduced an approach for active search of stiff inclusions such as tumors, arteries and other stiff inclusions in tissues by means of forceful palpation. We incorporated three different active learning objectives, namely active area search, active level sets and uncertainty sampling, into a stochastic trajectory optimization framework that respects the robot's kinematic and workspace constraints.  The results show that active area search algorithm performs better than active level sets, uncertainty sampling as well as the recently proposed Bayesian optimization-based methods that gained momentum in the literature. Additionally, our formulation enables incorporating uncertainty in robot position and force measurement. Accurate modelling of the interaction between the tissue and and continuum robots, and accurate force sensing remains to be significant challenges in this domain and will be a focus of future work. The future work will also focus on incorporating tissue mechanics models in stiffness estimation.

\bibliographystyle{IEEEtran}
\bibliography{references}
\end{document}